\documentclass[journal]{IEEEtran}
\ifCLASSINFOpdf
\else
\fi

\usepackage{amsmath}
\usepackage{algorithm}
\usepackage{algorithmicx}
\usepackage{algpseudocode}
\usepackage{float}
\usepackage{graphicx}
\usepackage{rotating} 
\usepackage{multirow}  
\usepackage{makecell}
\usepackage{amssymb}

\begin{document}
%
\title{Exposing the Copycat Problem of Imitation-based Planner: A Novel Closed-Loop Simulator, Causal Benchmark and Joint IL-RL Baseline}
%
%
%

\author{Hui~Zhou,
        Shaoshuai~Shi,
        Hongsheng Li 
\thanks{Hui Zhou and Hongsheng Li are with the Department of Electronic Engineering, The Chinese University of Hong Kong.}
\thanks{Shaoshuai Shi is with DiDi Global. This work was done during hui's internship at Didi Global.}
\thanks{Manuscript received April 19, 2005; revised August 26, 2015.}}

%
%

\markboth{Journal of \LaTeX\ Class Files,~Vol.~14, No.~8, August~2015}%
{Shell \MakeLowercase{\textit{et al.}}: Bare Demo of IEEEtran.cls for IEEE Journals}
%



\maketitle

\begin{abstract}

Machine learning (ML)-based planners have recently gained significant attention.  They offer advantages over traditional optimization-based planning algorithms.  These advantages include fewer manually selected parameters and faster development.  Within ML-based planning, imitation learning (IL) is a common algorithm.  It primarily learns driving policies directly from supervised trajectory data. While IL has demonstrated strong performance on many open-loop benchmarks, it remains challenging to determine if the learned policy truly understands fundamental driving principles, rather than simply extrapolating from the ego-vehicle's initial state. Several studies have identified this limitation and proposed algorithms to address it.  However, these methods often use original datasets for evaluation.  In these datasets, future trajectories are heavily dependent on initial conditions.  Furthermore, IL often overfits to the most common scenarios.  It struggles to generalize to rare or unseen situations.

To address these challenges, this work proposes: 1) a novel closed-loop simulator supporting both imitation and reinforcement learning, 2) a causal benchmark derived from the Waymo Open Dataset to rigorously assess the impact of the copycat problem, and 3) a novel framework integrating imitation learning and reinforcement learning to overcome the limitations of purely imitative approaches. The code for this work will be released soon.

\end{abstract}


%
\IEEEpeerreviewmaketitle

\section{Introduction}
%
%
%
%
\IEEEPARstart{A}{utonomous} driving has garnered significant attention in the research community, with a particular focus on Machine Learning-based Planning \cite{caesar2021nuplan, Dauner2023CORL} and End-to-End (E2E) learning systems \cite{Hu_2023_CVPR, chen2024end} recently. In contrast to the conventional modular approach—which segments the driving task into distinct components like perception, prediction, planning, and control—there is a growing trend towards learning-based methods. These methods leverage data to train systems that can potentially handle the complexities of driving in a more integrated and efficient manner. Although E2E methods hold great promise in the long run, they currently lack explainability and stability, requiring further experimentation and the development of additional testing tools, such as world simulators. The ML-planner offers a solid bridge between traditional methods and E2E approaches.

\begin{figure}[t]
  \centering
  \includegraphics[width=\columnwidth]{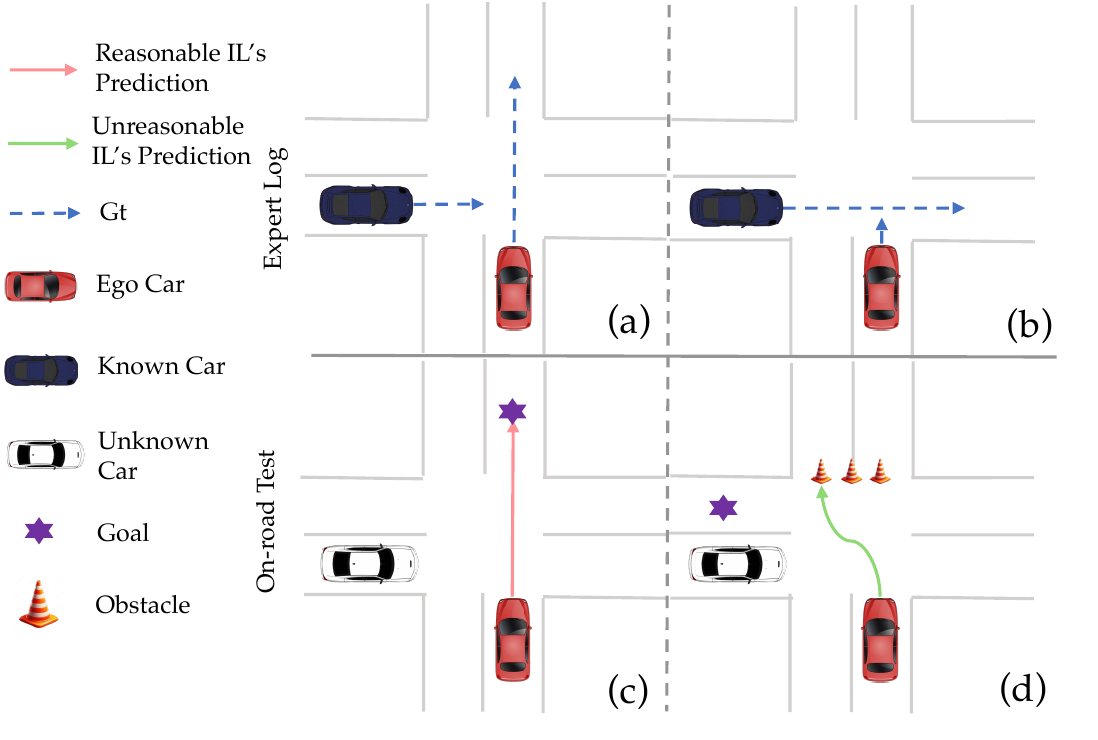}
  \caption{Illustration of the copycat problem. (a)-(b) At a certain intersection, the most frequently observed behavior is going straight. (c) Policies trained using imitation learning tend to perform well in scenarios that are close to the training data. However, when typical goals or behaviors are unavailable, the learned policy may generate unreasonable trajectories.}
  \label{fig:intro}
\end{figure}

In contrast to optimization-based planning methods \cite{ma2015efficient}, which rely heavily on manually set parameters, machine learning (ML) planners learn policies directly from data without requiring expert priors. However, ML-planners also have certain limitations. These include difficulties in adapting to new environments and shortcut learning, issues that are also prevalent in other deep learning approaches \cite{geirhos2020shortcut}. Recent study \cite{li2024ego} has increasingly emphasized the critical role of ego status in end-to-end systems, revealing that even a simple multilayer perceptron network (MLP) leveraging ego status can surpass state-of-the-art methods that rely on intricate modules. Some researchers have referred to this phenomenon as the copycat problem \cite{wen2020fighting, chuang2022resolving}, suggesting that such methods primarily predict a straightforward extrapolation from historical states. Furthermore, ML-Planner adopts a globally optimized approach over the limited training dataset, which can occasionally lead to unreasonable actions, as illustrated in Fig. \ref{fig:intro}.

In this paper, we focus on analyzing different driving actions in the same context but with varying endpoints, which may raise concerns about copycat behavior. First, we develop a novel closed-loop simulator to support simulation of different driving behaviors, along with a causality benchmark. The simulator includes a data loader, evaluation metrics, a basic simulation environment, policies for IL and RL algorithms, motion dynamics, an MPC controller, and visualization tools. Notably, our simulator is lightweight and does not rely on heavy GUI dependencies, making it easy to integrate with other state-of-the-art ML-planning algorithms. The causality benchmark is created by assigning different goals within the same contextual input (historical locations), mitigating the extrapolation effects of ego status used in ML-Planner. For each scenario, in addition to the original trajectory's goal, we generate additional future goals using Depth-First Search (DFS) with actions such as lane changing, left turning, right turning, or going straight. The benchmark is based on the Waymo Motion Prediction dataset \cite{waymo}, which is more challenging than nuPlan \cite{caesar2021nuplan}—a dataset commonly used in recent studies that typically evaluates around 1,118 scenarios, as referenced by PDM \cite{Dauner2023CORL}. This increased difficulty arises from the lack of routing paths\footnote{Routes are not released in WOMD.}. Second, given that some ML-Planners rely on waypoint prediction heads similar to traditional prediction methods, we use MTR \cite{shi2022motion, shi2024mtrplus} which has achieved top results on the Waymo motion prediction dataset, as our imitation learning baseline. Building on the previous methods, we introduce a new approach called MTR-SAC, which integrates Imitation Learning (IL) and Reinforcement Learning (RL) in the feature and action space. The RL component can learn without ground-truth data. We show that this method significantly improves the safety and adaptability of the learned policies compared to imitation-only methods, showing a clear improvement of IL-RL over a pure IL baseline.

The key contributions of our work are: (1) we introduce a new closed-loop simulator and causality benchmark based on a large real-world vehicle dataset, the Waymo Open Dataset, to facilitate causality evaluation in ML-Planners. (2) We propose a novel RL-IL approach and systematically evaluate its performance compared to baseline methods using our causality benchmark, demonstrating that combining imitation learning (IL) and reinforcement learning (RL) improves policy adaptability beyond what imitation learning alone can achieve.

\section{Related Works}

\subsection{Machine Learning based Planner}

Existing research in machine learning (ML)-based planning for autonomous driving can be divided into three categories: imitation learning, reinforcement learning, and hybrid approaches.

Imitation learning (IL) has become a promising method for training driving policies from expert demonstrations. Bojarski et al. \cite{bojarskiend} pioneer end-to-end autonomous driving by using convolutional neural networks to directly map raw pixel inputs to steering commands. To improve the robustness and versatility of IL models, Codevilla et al. \cite{codevilla2018end} propose conditional imitation learning, incorporating high-level commands to influence the learned policy. VAD \cite{jiang2023vad}, and UniAD \cite{Hu_2023_CVPR} both explore end-to-end approaches for IL, enabling agents to learn from large datasets.

Reinforcement learning (RL) offers the capability to learn optimal driving policies through interaction with the environment. Folkers et al. \cite{folkers2019controlling} train a neural network agent to map its estimated state to acceleration and steering commands by proximal policy optimization \cite{schulman2017proximal}. Wang et al. \cite{wang2018deep} demonstrate the application of deep reinforcement learning with the use of Deep Deterministic Policy Gradient algorithm to learn continuous control policies for autonomous driving. Tang et al. \cite{tang2022highway} solve the decision-making and planning problem with continuous action space approaches with Soft Actor Critic \cite{haarnoja2018soft} to improve the safety and efficiency of autonomous vehicles in highway scenarios.

Hybrid approaches combine planning methods from any two of the following categories: rule-based, imitation learning, and reinforcement learning. Dauner et al. \cite{Dauner2023CORL} develop a planner that uses the classical Intelligent Driver Model (IDM) \cite{treiber2000congested} for precise short-term actions, and integrate a learned ego-forecasting component for improved long-range accuracy. Lu et al. \cite{lu2023imitation} enhance robustness in challenging scenarios by using a weighted combination of imitation learning (IL) and reinforcement learning (RL) objectives. This is the first application of this combined approach in autonomous driving using substantial real-world human driving data. Huang et al. \cite{huang2022efficient} demonstrate the effectiveness of guiding an RL agent with an expert imitation policy, by regularizing the Kullback–Leibler divergence between the agent's policy and the expert's, specifically in unprotected left turns and roundabouts. Booher et al. \cite{booher2024cimrl} combines imitation learning with reinforcement learning by restricting the action space to an efficient support derived from the motion prior generated by a pre-trained imitation learning model.

\subsection{The Copycat Problem}

The copycat problem \cite{geirhos2020shortcut, li2024ego, wen2020fighting, chuang2022resolving}, within the context of machine learning, refers to situations where a model appears to perform well on a task but actually relies on unintended shortcuts or spurious correlations in the data, rather than learning the desired underlying concepts. This often leads to poor generalization when the model encounters data that differ slightly from its training distribution. The copycat problem represents a significant challenge in the field of imitation learning, where an agent must learn to replicate the behavior of a demonstrator. This problem is characterized by the need for the learner not only to mimic the actions of the demonstrator but also to understand and generalize the demonstrated behavior in varying contexts.

Recent studies often propose data augmentation strategies or add regularization terms to the optimization function to address this problem. For example, Bansal et al. \cite{bansal2018chauffeurnet} introduce a dropout mechanism on the past pose history for half of the examples, forcing the network to rely on other cues in the environment during training. Similarly, Seo et al. \cite{seo2024regularized} present a principled framework for behavior cloning using Past Action Leakage Regularization (PALR), which prevents the imitator from overfitting to leaked past information.

\subsection{Open-Loop and Closed-loop Simulators}

An open-loop system in autonomous driving refers to a setup where the vehicle's control actions are determined based on pre-set inputs without taking into account the current state or outputs of the system. This means that there is no accumulation of error or compounding effects in real time. For example, some waypoints predicted by imitation learning may not be feasible for real vehicles to follow because they can violate hardware constraints, such as exceeding the maximum acceleration value. In the closed-loop system, the vehicle continuously monitors its environment through various vehicle models and adjusts its actions based on real-time feedback. This could involve steering adjustments, acceleration, braking, or lanekeeping. The key advantage of a closed-loop system is its ability to react dynamically to changes in real time.

Recent studies, such as nuPlan \cite{caesar2021nuplan}, CARLA \cite{dupuis2010opendrive}, Waymax \cite{gulino2024waymax}, SMARTS \cite{zhou2021smarts}, and MetaDrive \cite{li2022metadrive}, focus on closed-loop simulators due to their closer resemblance to real-world vehicle behavior compared to open-loop simulators. Among these simulators, CARLA and MetaDrive offer rendering capabilities and can provide virtual sensor data, while nuPlan and Waymax support real vehicle datasets.

\section{Methods}

The proposed method consists of three parts. The first part involves building a closed-loop simulator with the following characteristics: It should include basic policies and control strategies for all moving objects. For reinforcement learning training, it also needs to have a basic reward calculator and core algorithms. Notably, parallel simulation is a crucial feature. The second part involves constructing a causal dataset, aiming at simulating the response to different endpoints under the same input conditions, and measuring the impact of the copycat problem. The third part introduces a joint training framework that combines imitation learning and reinforcement learning, designed to mitigate the effects of the copycat problem.

\begin{figure*}[t]
  \centering
  \includegraphics[width=\textwidth]{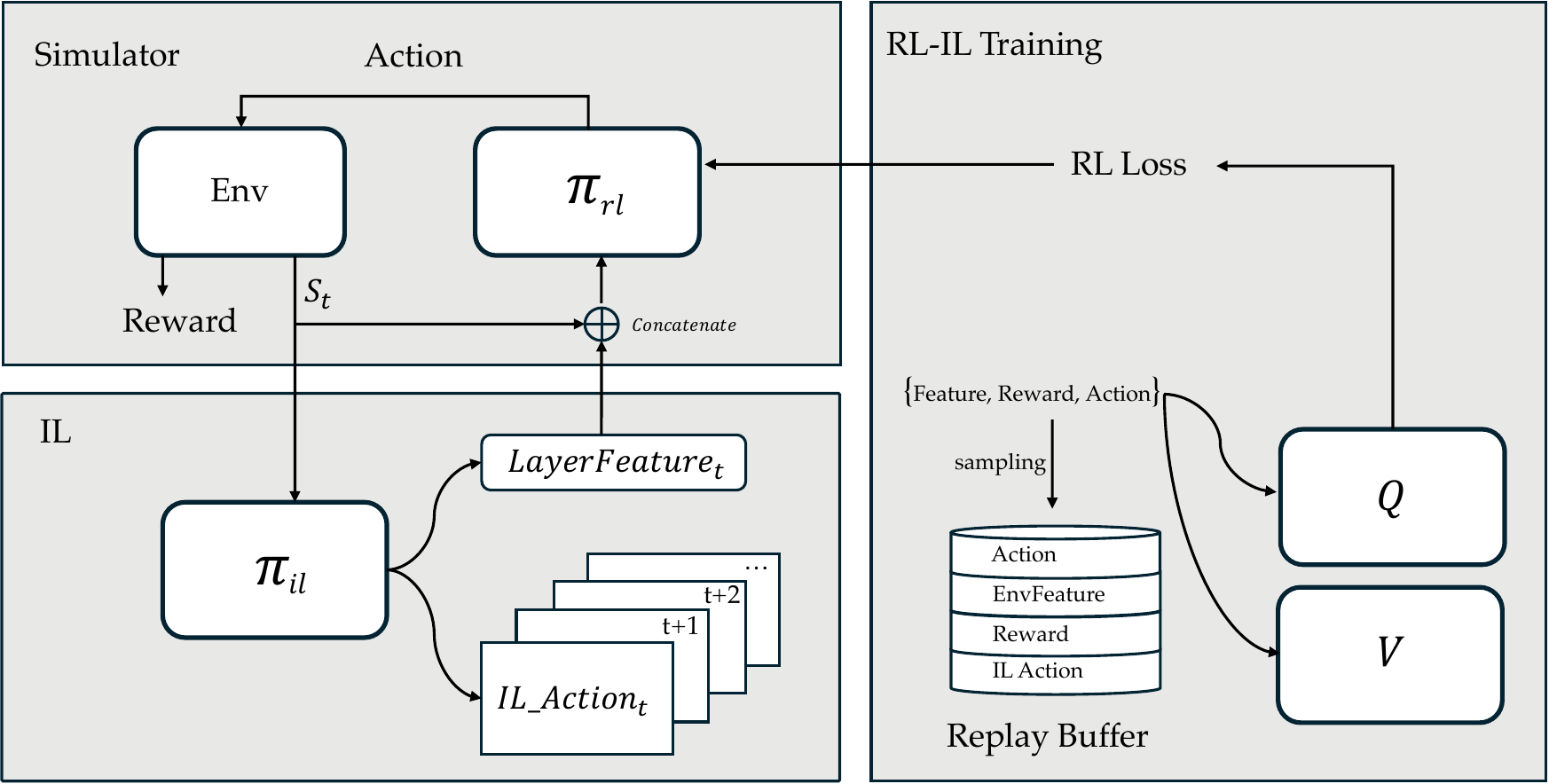}
  \caption{The framework is divided into three parts. The first part is the closed-loop simulator, which initializes the environment, executes actions, and outputs the resulting state and reward. The second part focuses on Imitation Learning (IL), utilizing a pre-trained MTR \cite{shi2022motion}. This part takes the current state from the simulator as input and produces transformer-encoded features along with future trajectories. The third module involves reinforcement learning, specifically employing the Soft Actor-Critic (SAC) algorithm for offline learning.}
  \label{fig:framework}
\end{figure*}

\subsection{The Closed-Loop Simulator}

 The Waymo Open Dataset \cite{waymo} can offer more scenarios compared to other datasets. Currently, several simulators support the Waymo dataset, including InterSim \cite{sun2022intersim}, TBSim \cite{xu2023bits}, Tactics2d \cite{li2023tactics2d}, and Waymax \cite{gulino2024waymax}. InterSim provides a good baseline, but lacks multi-GPU support. TBSim is an excellent open-source library, but it relies on trajdata. We find that trajdata's fundamental data structure has a drawback that prevents it from fully loading Waymo formatted data\footnote{https://github.com/NVlabs/trajdata/issues/36}. Tactics2d currently does not release their urban environment. Waymax is built using JAX, which poses a significant integration challenge for existing PyTorch-based imitation learning algorithms. For instance, gradient backpropagation is difficult to translate from JAX's computational graph to PyTorch's computational graph. To address these issues, we developed a new closed-loop simulator based on PyTorch, providing support for distributed training and evaluation on multiple GPUs. Due to Waymax's professional design, our simulator adopts many similar features. The main objective of our simulator is to develop a lightweight framework that supports the Waymo dataset and can simultaneously support closed-loop simulation for both imitation learning and reinforcement learning training.

Our simulator consists of the following modules:
\begin{itemize}
    \item Dataloader: data loading.
    \item Policy: \begin{itemize}
              \item \textbf{MTR} \cite{shi2022motion}: IL-based planner.
              \item \textbf{StateSAC}: RL-based planner.
              \item \textbf{MTR-SAC}: joint IL-RL planner.
              \item \textbf{IDM} \cite{treiber2000congested}: Intelligent Driver Model.
              \item \textbf{Expert}: replay of the original vehicle or pedestrian actions\end{itemize}
    \item RLCore:
        \begin{itemize}
            \item \textbf{SAC} \cite{haarnoja2018soft}: offline RL.
        \end{itemize}
    \item Dynamics:
        \begin{itemize}
            \item \textbf{Default}: using the planner prediction as the position in the next frame directly.
            \item \textbf{Bicycle} \cite{polack2017kinematic}: the next position is recalculated using the bicycle motion model based on the planner's prediction, which may introduce accumulated errors compared to the original model output.
        \end{itemize}
    \item Metrics (Rewards):
        \begin{itemize}
          \item \textbf{Offroad}: crossing over the road edge or solid line.
          \item \textbf{Collision}: checking the intersection of bounding boxes.
          \item \textbf{Uncomfort}: exceeding the maximum acceleration value.
          \item \textbf{Completion}: outputting done when distance to the goal is less than a specified threshold.
          \item \textbf{Progress}: the difference in the distance to the endpoint between the current and previous frame (only used in RL training) \end{itemize}
    \item Environment: managing the base environment and scheduling different modules.
    \item Controller: using the Model Predictive Control (MPC) to follow the predicted waypoints.
    \item Visualization: visualizing all objects and maps from the bird view.
\end{itemize}

Our simulator supports both reactive (IDM, MTR, and your custom algorithm) and non-reactive closed-loop simulations (using Expert) for NPC and SDC vehicles. For online reinforcement learning, we find that when using multiple GPUs for simulation, each scenario may not finish at the same time. For example, the policy might take longer to run in some easy scenarios compared to harder ones. To address this issue, we adopt the offline RL algorithm as the baseline and always simulate a large batch of scenarios on a single gpu and store all states in a replay buffer. Once all GPUs have completed their simulations, we begin training the reinforcement learning algorithm. After completing the RL training, we continue to run the simulation on each GPU.

\subsection{Causality Benchmark}

Previous works \cite{li2024ego, chuang2022resolving, montali2024waymo} identify the copycat problem in machine learning-based planning and propose methods to address it. However, these evaluations occur on the original dataset where future actions are highly correlated with past actions. The Waymo team also recognizes this issue and introduces a new benchmark to evaluate the causality of agent behavior \cite{sun2024causalagents}. Their approach involves randomly removing some agents, with less focus on the driving actions of self-driving-car (SDC) in a closed-loop simulator. In our paper, we aim to assess this issue using an extensive benchmark.

Generating endpoints is difficult because the Waymo dataset has a different format than other datasets. Datasets like nuPlan \cite{caesar2021nuplan} use a map format similar to the OpenDRIVE format, which includes features such as road blocks, road segments, and others. Waymo's format makes it hard to use some open-source routing algorithms. Inspired by instruction navigation, we propose a simple strategy to generate various destinations from 1-second trajectory input, based on lane exits and neighbors\footnote{https://github.com/waymo-research/waymo-open-dataset/blob/master/docs}, with four actions: changing lane, turning left, turning right, or going straight. We use a Depth-First Search (DFS) with a cost to avoid unrealistic actions, like changing two lanes at once. After generating several candidate paths, we will select the point on each path that is equidistant from the starting point as the corresponding point on the original trajectory. Finally, we use an NMS method to remove some nearest endpoints generated by different action sequences. 

\begin{algorithm}
  \caption{Causality Benchmark}
  \label{alg:causality_benchmark}
  \begin{algorithmic}[1]
    \State \textbf{Predefined Parameters:}
    \State \hspace*{\algorithmicindent} $CostTable \gets \{\text{LC}: 5, \text{TL}: 1, \text{TR}: 1, \text{Go}: 1\}$
    \State \hspace*{\algorithmicindent} $MaxCostThres \gets 10$
    \State \hspace*{\algorithmicindent} $NmsThres \gets 2.5$ meter
    \State \textbf{Input:} location $\textit{s}$ at 1-second, and $\textit{s2g\_dist}$
    \State \textbf{Output:}  multiple goals $\textit{m}$
    \State $\textit{path} \gets \emptyset$
    \Procedure{SearchPart}{$\textit{curr\_cost}$, $\textit{curr\_path}$}
        \If {$\text{MaxDist}(\textit{curr\_path}) > \textit{s2g\_dist}$}
            \If {$\textit{curr\_cost} \leq \textit{MaxCostThres}$}
                \State $\textit{path} \gets \textit{path} + [\textit{curr\_path}]$
            \EndIf
            \State \Return
        \EndIf
        \For {each $\textit{action}$ in $\textit{actions}$}
            \State $\textit{new\_cost} \gets \textit{curr\_cost} + \textit{CostTable}[\textit{action}]$
            \State $\textit{new\_path} \gets \textit{curr\_path} + [\textit{lane\_points}]$
            \State \Call{SearchPart}{$\textit{new\_cost}$, $\textit{new\_path}$}
        \EndFor
    \EndProcedure
    \State \Call{SearchPart}{$0$, $\textit{s}$} 
    \State $\textit{candidates} \gets \text{FindGoalsByDist}(\textit{path}, \textit{s2g\_dist})$
    \State $\textit{m} \gets \text{GoalsNMS}(\textit{candidates}, \textit{NmsThres})$
    \State \Return $\textit{m}$
  \end{algorithmic}
\end{algorithm}

\subsection{Integrated Imitation and Reinforcement Learning}

We present a strong baseline to mitigate the inherent drawbacks of imitation learning (IL), particularly its challenges in learning complex behaviors in scenarios with limited data due to data distribution mismatches. We leverage reinforcement learning (RL) to address these challenges. To establish a baseline, we initially trained MTR \cite{shi2022motion} algorithm on the Waymo Open dataset and develop a simple method for integrating IL and RL: concatenates learned features from IL with the simulator's state representation—comprising location, velocity, and the endpoint—to construct the environmental observation for the RL agent. The complete pipeline is illustrated in Fig. \ref{fig:framework}.


\subsubsection{MTR}

The architecture of MTR is based on a transformer encoder-decoder framework. The encoder processes the historical trajectories of the ego-vehicle and surrounding agents, along with relevant map information. The decoder then generates multiple possible future trajectories. Extensive experiments conducted on benchmark datasets demonstrate that MTR achieves superior performance compared to existing methods, particularly on the Waymo Open Dataset. Specifically, it exhibits higher precision in trajectory prediction and demonstrates improved robustness in handling rare and challenging scenarios. Moreover, the modular design of MTR facilitates seamless integration with other components of autonomous driving systems, making it a strong baseline for motion prediction tasks.

\subsubsection{RL Training with IL Feature}

In our experiments, we use the feature extracted from the layer immediately preceding the motion prediction head of the MTR model as our imitation observation feature. This choice is motivated by the fact that MTR uses 64 intention queries in diverse directions, effectively capturing a wide range of potential paths. For the reinforcement learning component, we adopt the Soft Actor-Critic (SAC) algorithm \cite{haarnoja2018soft} as our baseline due to its demonstrated stability and superior performance across various tasks. Since we are working with a real vehicle dataset, we train the RL model in a continuous action space, where the policy predicts both the mean and variance of a Gaussian distribution to generate acceleration and turning rate commands. The policy in SAC algorithm with maximum entropy regularization is formulated as follows:

\[
\pi^* = \arg\max_{\pi} \mathbb{E}_{\tau \sim \pi} \left[ \sum_{t=0}^{\infty} \gamma^t \left( R(s_t, a_t, s_{t+1}) + \alpha H\left(\pi(\cdot | s_t)\right) \right) \right]
\]

The loss function is:

\[
L_{\pi}(\theta) = \mathbb{E}_{s_t \sim R, a_t \sim \pi_{\theta}} \left[ \alpha \log(\pi_{\theta}(a_t | s_t)) - Q(s_t, a_t) \right]
\]

The specific reinforcement learning reward function is composed of the following five terms:

\begin{itemize}
    \item \textbf{Collision}: \( R_{\text{collision}} = \min(d_{\text{collision}} - 1.0, 0) \)
    \item \textbf{Offroad}: \( R_{\text{offroad}} = \mathrm{clip}(-1.0 - d_{\text{offroad}}, -2, 0) \)
    \item \textbf{Progress}: \( R_{\text{progress}} = \mathrm{clip}(d_{\text{progress}} - 0.1, -2, 1) \)
    \item \textbf{Smoothness}: \( R_{\text{smoothness}} = -0.5 * \mathrm{logit\_or} ( \Delta_{\text{accel} } > 1.5, \Delta_{\text{turning} } > 0.1) \)
    \item \textbf{Completion}: \( R_{\text{completion}} = \mathrm{is\_goal} \, ? \, 10 :0 \)
\end{itemize} \(d_{\text{collision}}\) is the euclidean distance in meters between the closest points of the ego vehicle and the nearest bounding box of other vehicles. \(d_{\text{offroad}}\) is the distance in meters from the vehicle to the nearest road edge (negative values indicate on-road, positive values indicate off-road). \(d_{\text{progress}}\) is calculated as the difference between the distance to the goal at time \(t - 1\) and the distance to the goal at time \(t\). \(\Delta_{\text{accel} }\) or \(\Delta_{\text{turning} }\)  is calculated as the difference between the current action at time \(t\) and the previous action at time \(t-1\). Our final reward is the sum of these terms. This reward structure encourages the agent to learn safe and efficient navigation while maintaining smooth and controlled driving behavior.

\section{Experiments}

Our framework involves training two models: one for imitation learning and another one for joint imitation and reinforcement learning.

\subsection{Training Dataset}

\subsubsection{Imitation Learning}

as mentioned in MTR \cite{shi2022motion}, the model is trained on the Waymo Open Motion Dataset (WOMD), which provides 1 second of history data and aims to predict 6 marginal or joint trajectories of the agents for 8 seconds into the future. The dataset consists of approximately 487,000 training scenes, 44,000 validation scenes, and 44,000 testing scenes.

\subsubsection{Reinforcement Learning}

In the standard setting of WOMD, the dataset provides a set of agent indices for prediction. However, we find that in most scenarios, the index of the self-driving car (ego car) is not included in this set. Therefore, we filter out these scenarios from our training data and only keep the ones that include the ego car index, resulting in approximately 50,000 scenarios. It is worth noting that during the training stage, to maintain consistency with real vehicle data, we do not use data augmentation to generate different goals.
\subsection{Causality Test Dataset}

\begin{figure*}[t]
  \centering
  \includegraphics[width=\textwidth]{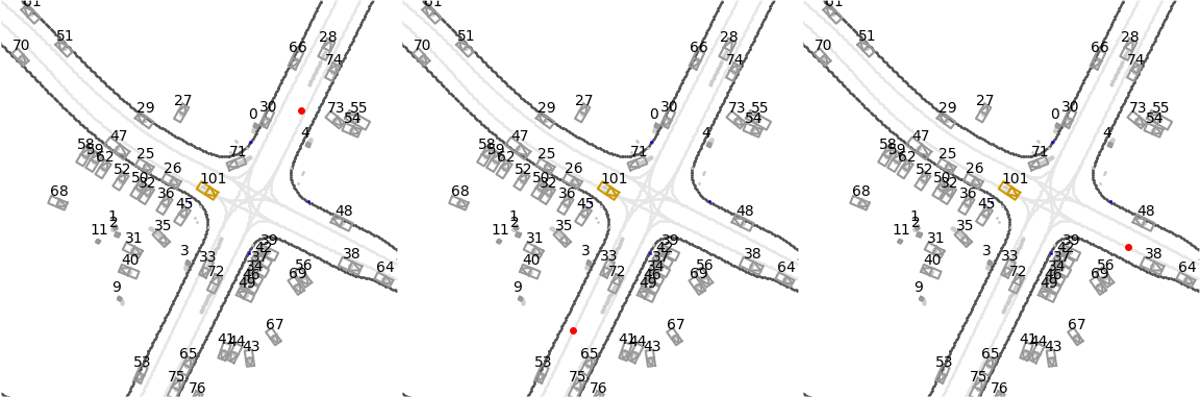}
  \caption{One typical example in our causality benchmark illustrates going straight, turning right, and turning left. The red point is the goal.}
  \label{fig:causality_benchmark}
\end{figure*}

We apply the same filtering strategy to the test scenarios, removing those that do not include the ego car index. We also remove scenarios where the ego car is static, resulting in 3,888 remaining scenarios, referred to as \textbf{Test4k}. After applying our augmentation method, the number of scenarios increases to approximately 9,096, named \textbf{Causality9k}. Some visualizations can be found in Fig.\ref{fig:causality_benchmark}.

\subsection{Training Details}

For MTR training, we follow the default settings in the open-source version using all available data\footnote{https://github.com/sshaoshuai/MTR}. To efficiently conduct an ablation study for joint IL-RL training, we uniformly sample 20\% of scenarios (10k scenes) from the RL training dataset in their default order and empirically find that this subset has a distribution similar to the full training set. All models are evaluated using the same metric on the causality benchmark.

For joint IL and RL training, we train the model for a maximum of 10 epochs using 8 A6000 GPUs for the ablation study, which takes approximately 4 days, respectively. We use the Adam optimizer with a learning rate of 0.0001 and set the tau parameter to 0.005. The batch size is set to 1 per GPU. For RL training, we update the model after every 32 scenarios, with each GPU simulating four scenarios and saving all states into the replay buffer. We do not explore many sophisticated architectures for RL training. Instead, we use simple MLPs with 6 to 10 layers for our two Q networks, the value (V) network, and the actor network.

\subsection{Results}

We evaluate the performance of different models based on four metrics: completion, collision, offroad, and stuck. The completion metric is assessed by calculating the distance from the current location to the target goal. A SDC is considered complete if it achieves a distance of less than 2 meters. The collision and off-road metrics are determined by examining bounding box overlap, where overlap signifies a collision with another object or an excursion beyond the road edge, respectively. If a vehicle does not reach its goal, experiences a collision, or goes off-road, it is currently classified as stuck. We are working on a routing algorithm that will measure the progress of stuck vehicles using routing points; however, this functionality is not included in the current simuator to support Waymo dataset.

\subsubsection{Open-Loop and Closed-Loop Evaluation}

To verify our simulator, we designed two evaluations: one for open-loop and another for closed-loop. For the open-loop evaluation, we only use MTR for a single prediction, and the following steps are sampled from the prediction by index. In the closed-loop approach, we do not use the imitation learning prediction directly because we find that the location at the next step (0.1 second) is not stable. Therefore, we first apply a Model Predictive Control (MPC) algorithm to compute the required actions for the next 1-second based on the predicted trajectory for global optimization.

As shown in Table \ref{table:tb1}, there is a significant performance difference between MTR in open-loop and closed-loop configurations. Furthermore, performance decreases considerably without an MPC controller to refine the original MTR prediction.

\begin{table}[h]
\begin{tabular}{|l|c|c|c|c|c|c|c|}
\hline
& SDC & Dyn. & NPC & Compl. & \multicolumn{3}{c|}{UnCompl. \textbf{$\downarrow$}} \\
\cline{6-8}
\multirow{5}{*}{\rotatebox[origin=c]{90}{Test4k}} & & & & \textbf{$\uparrow$} & Col. & Off. & Stu. \\
\cline{2-8}
& Expert & Default & Expert & 0.998 & 0.001 & 0.001 & 0.0 \\
\cline{2-8}
& Expert & Default & IDM & 0.823 & 0.176 & 0.001 & 0.0 \\
\cline{2-8}
& MTR-Open & Default & Expert & 0.474 & 0.065 & 0.188 & 0.273 \\
\cline{2-8}
& MTR-Open & Default & IDM & 0.379 & 0.216 & 0.180 & 0.225 \\
\cline{2-8}
& \makecell{MTR-Close \\ w/o MPC} & Default & IDM & 0.154 & 0.163 & 0.160 & 0.53 \\
\cline{2-8}
& \makecell{MTR-Close \\ w MPC} & Default & IDM & 0.513 & 0.124 & 0.042 & 0.321 \\
\cline{2-8}
& \makecell{MTR-Close \\ w/o MPC} & Bicycle & IDM & 0.093 & 0.177 & 0.397 & 0.333 \\
\cline{2-8}
& \makecell{MTR-Close \\ w MPC} & Bicycle & IDM & 0.510 & 0.122 & 0.041 & 0.327 \\
\hline
\end{tabular}
\caption{The results of MTR for both open-loop and closed-loop on the Test4k dataset.}
\label{table:tb1}
\end{table}

\subsubsection{Causality Benchmark}

For the causality benchmark, it is unfair to compare the original MTR, which uses the trajectory with the maximum probability out of six as the prediction. Therefore, in this benchmark, we use the trajectory nearest to the goal as the policy at each step, 
called \(MTR\_Close^*\).

Compared with Table \ref{table:tb1} and Table \ref{table:tb2}, the results show that our Causality9k is more diverse and difficult than the original dataset.

\begin{table}[h]
\begin{tabular}{|l|c|c|c|c|c|c|c|}
\hline
& SDC & Dyn. & NPC & Compl. & \multicolumn{3}{c|}{UnCompl. \textbf{$\downarrow$}} \\
\cline{6-8}
\multirow{5}{*}{\rotatebox[origin=c]{90}{Causality9k}} & & & & \textbf{$\uparrow$} & Col. & Off. & Stu. \\
\cline{2-8}
& MTR-Close & Default & IDM & 0.205 & 0.141 & 0.046 & 0.608 \\
\cline{2-8}
& MTR-Close & Bicycle & IDM & 0.206 & 0.138 & 0.043 & 0.613 \\
\cline{2-8}
& MTR-Close* & Default & IDM & 0.308 & 0.155 & 0.09 & 0.447 \\
\cline{2-8}
& MTR-Close* & Bicycle & IDM & 0.308 & 0.156 & 0.089 & 0.447 \\
\hline
\end{tabular}
\caption{Closed-loop evaluation comparsion on our Causality9k benchmark}
\label{table:tb2}
\end{table}

\subsubsection{Combing IL with RL}

Building upon our imitation-only baseline, we propose two additional experiments: a reinforcement learning-only approach and an integrated imitation learning-reinforcement learning (IL-RL) framework. An ablation study, detailed in Table \ref{table:tb3} and \ref{table:tb4}, revealed that our simplest IL-RL fusion method provides a robust baseline for combining imitation learning and reinforcement learning. In our experiments, vehicle dynamics are modeled using a bicycle model, and NPCs are controlled using the Intelligent Driver Model (IDM). The \(StateSAC\) method uses the original map's polylines, detected objects, the designated goal, and the SDC's current state as environmental features during reinforcement learning training. For a fair comparison, \(StateSAC\) is also trained for 10 epochs.

\begin{table}[h]
\begin{tabular}{|l|c|c|c|c|c|c|}
\hline
& SDC & Strategy & Compl. & \multicolumn{3}{c|}{UnCompl. \textbf{$\downarrow$}} \\
\cline{5-7}
\multirow{5}{*}{\rotatebox[origin=c]{90}{Test4k}} & & & \textbf{$\uparrow$} & Col. & Off. & Stu. \\
\cline{2-7}
& MTR-Close & IL only & 0.510 & 0.122 & 0.041 & 0.327 \\
\cline{2-7}
& StateSAC & RL only & 0.044 & 0.339 & 0.560 & 0.057 \\
\cline{2-7}
& MTR\_SAC & RL\_IL & 0.580 & 0.273 & 0.142 & 0.005 \\
\hline
\end{tabular}
\caption{Closed-loop evaluation comparison between pure IL, RL and IL-RL methods on Test4k benchmark.}
\label{table:tb3}
\end{table}

\begin{table}[h]
\begin{tabular}{|l|c|c|c|c|c|c|}
\hline
\multirow{5}{*}{\rotatebox[origin=c]{90}{Causality9k}} & SDC & Strategy & Compl. & \multicolumn{3}{c|}{UnCompl. \textbf{$\downarrow$}} \\
\cline{5-7}
 & & & \textbf{$\uparrow$} & Col. & Off. & Stu. \\
\cline{2-7}
& MTR-Close* & IL only & 0.308 & 0.156 & 0.089 & 0.447 \\
\cline{2-7}
& StateSAC & RL only & 0.041 & 0.328 & 0.563 & 0.068 \\
\cline{2-7}
& MTR\_SAC & RL\_IL & 0.496 & 0.223 & 0.268 & 0.013 \\
\hline
\end{tabular}
\caption{Closed-loop evaluation comparison between pure IL, RL and IL-RL methods on our Causality9k benchmark.}
\label{table:tb4}
\end{table}

\subsubsection{Reward Weights}

In RL training, the balance between safety and progress is controlled by the reward weights. For completion and smoothness, we keep their weights fixed. Specifically, we use weights, such as $(1, 1, 1)$ in Table. \ref{table:tb5}, to represent the relative importance of off-road, collision, and progress rate, respectively.

\begin{table}[h]
\begin{tabular}{|c|c|c|c|c|c|}
\hline
Method & Reward Weight & Compl. & \multicolumn{3}{c|}{UnCompl. \textbf{$\downarrow$}} \\
\cline{4-6}
 & & \textbf{$\uparrow$} & Col. & Off. & Stu. \\
\cline{1-6}
\makecell{MTR\_SAC} & 1,1,1 & 0.486 & 0.230 & 0.268 & 0.016 \\
\cline{2-6}
Causality9k & 5,5,1 & 0.487 & 0.273 & 0.239 & 0.001 \\
\cline{2-6}
& 1,1,10 & 0.496 & 0.223 & 0.268 & 0.013 \\
\hline
\end{tabular}
\caption{Ablation study about reward weights on Causality9k.}
\label{table:tb5}
\end{table}







\subsection{Visualizations}







\begin{figure*}[t]
  \centering
  \includegraphics[width=\textwidth]{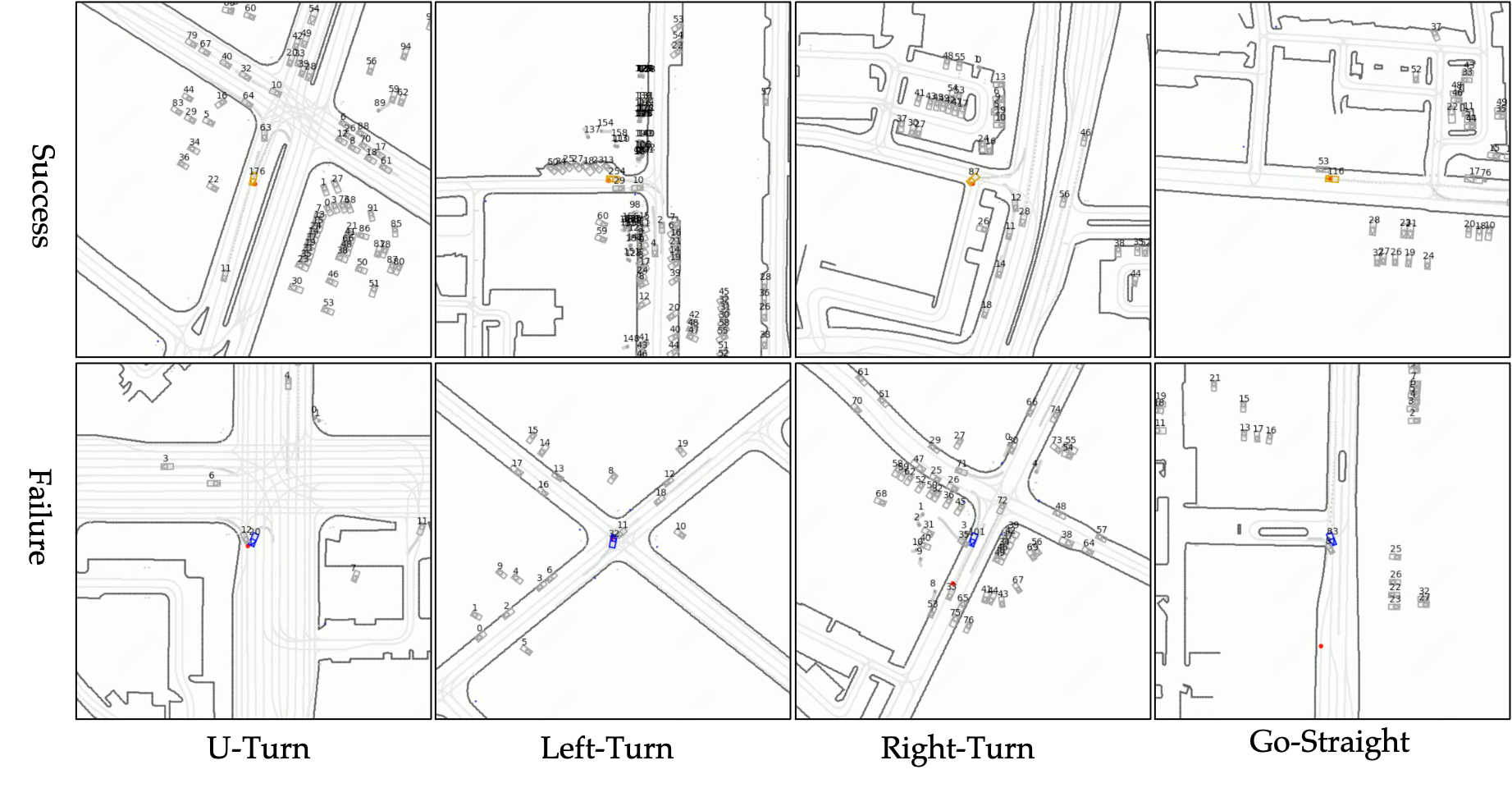}
  \caption{Success and failure cases in four scenarios: U-Turn, Left-Turn, Right-Turn, and Go-Straight.}
  \label{fig:failure}
\end{figure*}

We visualize some success and failure cases on the Causality9k benchmark in Fig. \ref{fig:failure} to aid future research, and find that the current model does not handle agent interactions perfectly.

\section{Conclusion}

In this paper, we present a novel framework to address the copycat problem in machine-learning-based planners. To validate our approach, we first design a causality benchmark that evaluates performance by altering goal conditions while keeping the input consistent. This benchmark eliminates the influence of historical actions, such as ego state dependencies, which are commonly used in most planning methods. To tackle this problem, we propose a new simulator that supports both imitation learning (IL) and reinforcement learning (RL) training. Through experiments, we demonstrate the effectiveness of a simple IL-RL baseline, showcasing its potential as a robust solution.

\section{Limitations}
In our framework, there is still much to explore. For example, designing an effective fusion strategy remains an open question. Although we propose a simple strategy, our results indicate significant potential for further investigation. Moreover, methods such as MTR and its variants demonstrate strong performance on the WOMD and Sim Agents benchmarks. However, based on our experiments, we believe that certain aspects are still overlooked, such as the physical constraints of vehicles.

\section*{Acknowledgment}

Specifically, we would like to thank the Waymax team for their open-source software, from which we have learned a lot to build our simulator.

\ifCLASSOPTIONcaptionsoff
  \newpage
\fi




\bibliographystyle{IEEEtran} 
\bibliography{references}

%








\end{document}